\documentclass[runningheads]{llncs}
\usepackage[T1]{fontenc}
\usepackage{graphicx}
\usepackage{hyperref}
\usepackage{color}

\usepackage{adjustbox}
\usepackage{array}
\usepackage{booktabs}
\usepackage{cite}
\usepackage{float}
\usepackage{latexsym}
\usepackage[table]{xcolor}

\makeatother

\newcommand{\comment}[1]{}

\begin{document}
\title{Redakto – The Incognito Tab for LLMs}
% \title{[Anonymous System] – The Incognito Tab for LLMs}
%
\titlerunning{Redakto – The Incognito Tab for LLMs}
\author{Saurav Kumar Saha\inst{1}\orcidID{0009-0002-9812-4150} \and
Tom Röhr\inst{1}\orcidID{0009-0007-7448-8941} \and
Felix Bießmann\inst{1,2}\orcidID{0000-0002-3422-1026}}

% \author{Anonymous Author(s)}

%
\authorrunning{S. K. Saha et al.}
% \authorrunning{Anonymous Author et al.}
\institute{Berlin University of Applied Sciences, Luxemburger Str. 10, 13353 Berlin, Germany\\
\email{\{sauravkumar.saha,tom.roehr,felix.biessmann\}@bht-berlin.de}\\
\url{https://www.bht-berlin.de}\and
Einstein Center Digital Future, Wilhelmstraße 67, 10117 Berlin, Germany\\
\url{https://www.digital-future.berlin}}

% \institute{Anonymous Institute(s)\\
% \email{Anonymous Email(s)}}

%
\maketitle              % typeset the header of the contribution
\begin{abstract}
Large Language Models (LLMs) are being increasingly used in everyday applications. A major challenge in the context of LLMs or Artificial Intelligence (AI) in general is to ensure privacy when using them, meaning that personally identifiable information (PII) is removed from any text that enters an LLM. These challenges have become more urgent with novel EU legislation. Uncertainty around LLM usage with respect to privacy concerns in EU countries can be a major blocker for the speed of innovation and transfer from research to applications. Here we present \textbf{Redakto}, a tool that can be used for anonymizing text prior to feeding it to an LLM or other downstream text processing. We provide state-of-the-art functionalities for both redaction of PII but also when used for pseudonymization. These functionalities are exposed such that they can easily be used by end-users, through the Redakto web application, and by developers and researchers, via REST APIs and model context protocol (MCP) hooks. The implementation is fully open source, requires modest compute resources, and can be readily deployed on local hardware.
In contrast to prior work and in order to better assess the quality of the anonymized texts, we conduct extensive empirical evaluations on textual data from legal and medical domain with respect to both privacy and utility of the redacted texts. Our empirical results demonstrate that the texts anonymized with different redaction strategies achieve utility scores on par with the original texts, suggesting that anonymization with Redakto can be used for LLM tasks without substantial negative impact for the tasks we explored.
% A video demonstration of Redakto can be found here, \url{https://cloud.bht-berlin.de/index.php/s/94mysedTHsR5dSq}

\end{abstract}
\section{Introduction}
In an era where large language models (LLMs) are transforming how we process and analyze text, the need to safeguard sensitive information has never been more critical.
While these models bear potential for research and applications, the risk of inadvertently exposing personally identifiable information (PII) remains a significant barrier to their safe and ethical use.
This challenge is particularly acute in fields such as public administration, healthcare, or legal studies, where the analysis of confidential documents is routine.
To foster more responsible usage of LLMs in these domains and beyond, we present a novel software demonstrator designed to automatically detect and redact PII from text, enabling users to confidently share sanitized content with LLM-based chatbots or other downstream applications where privacy is important.
Extending prior work on practical implementations of anonymizations, we develop evaluation protocols to evaluate the effectiveness of different redaction strategies such as semantic label masking, random masking etc. with respect to utility. Our experimental validations on a variety of textual tasks demonstrate that our solution not only preserves privacy but can also be used to maintain the integrity and utility of the original text, paving the way for secure and responsible AI-assisted analysis.
In summary, this study makes the following contributions:

\begin{itemize}
\item \textbf{Redakto} web application and APIs (REST \& MCP)
\item Privacy evaluation of Redakto models
\item Utility evaluation of redacted texts in downstream tasks
\end{itemize}

\section{\textbf{Redakto} web application and APIs}
The Redakto models are exposed in three different ways to make them as easy to use as possible for end users, developers, and researchers.
More concretely, we provide a readily usable web application as well as REST APIs and a Model Context Protocol (MCP) server which are hosted on institutional computing infrastructure.
All components are containerized and can be self-hosted (e.g., on Kubernetes clusters) and the MCP server can be integrated within MCP-compatible LLM assistants and agent frameworks, enabling privacy-preserving redaction as a native tool inside automated LLM workflows instead of just prompting. A brief overview of the system design and implementation architecture of Redakto is provided in Appendix~\ref{appendix:redakto_system_design}.

% \begin{figure}[h]
%    \centering
%     \begin{minipage}[t]{0.47\columnwidth}
%         \centering
%         \adjustbox{frame=0.5pt, clip}{
%             \includegraphics[width=\linewidth]{pii_detection_with_redakto.png}
%         }
%         {\footnotesize PII detection and removal}
%     \end{minipage}\hspace{0.04\columnwidth}%
%     \begin{minipage}[t]{0.47\columnwidth}
%         \centering
%         \adjustbox{frame=0.5pt, clip}{
%             \includegraphics[width=\linewidth{pseudonymization_with_redakto.png}
%         }
%         {\footnotesize Pseudonymization}
%     \end{minipage}
%     \caption{Screenshots of the Redakto web application}
%     \label{fig:screenshots_of_redakto_app}
% \end{figure}

\subsection{Web Application}
A publicly accessible instance of the Redakto web application is available at \url{https://redakto.demo.calgo-lab.de}.
% In \autoref{fig:screenshots_of_redakto_app} we highlight the web application when used for both {\em redaction} of PII but also when used for {\em pseudonymization}.
For ease of installation and deployment, we also make Redakto available as a Docker container image, accompanied by deployment documentation\footnote{\url{https://hub.docker.com/r/sksdotsauravs/redakto-app}}.

\subsection{REST APIs}
The REST API endpoints are documented and exposed with Swagger UI within the Redakto web application\footnote{\url{https://redakto.demo.calgo-lab.de/api/docs}}.
The first main endpoint
%, POST \texttt{/api/predict/detect\_entities}, 
performs named entity recognition over one or more input texts while the second one
%, POST \texttt{/api/predict/detect\_entities\_and\_pseudonymize}, 
extends this functionality by additionally generating pseudonymized versions of the input texts.
%A separate configurable parameter allows multiple distinct pseudonymized variants to be generated per document, which is useful for data augmentation and robustness evaluation.
Both endpoints support coarse- and fine-grained entity labeling.

% \begin{figure}
%     \centering
%     \includegraphics[width=\columnwidth]{text-redaction-with-mcp-and-llm-h.png}
%     \caption{\centering Screenshots of Redakto MCP server usage with Claude Desktop}
%     \label{fig:redakto-mcp-with-claude}
% \end{figure}

\subsection{MCP Server}
We additionally provide an MCP (Model Context Protocol) server, distributed as an npm package\footnote{\url{https://www.npmjs.com/package/@sksdotsauravs/redakto-app-mcp-server}}, 
that exposes Redakto’s functionalities as structured tools for LLM assistants and agent frameworks. Through this interface, LLM assistants such as Claude
%\autoref{fig:redakto-mcp-with-claude}) 
can invoke schema-defined tool calls for entity detection and pseudonymization, instead of relying on prompt-only processing. This enables privacy-aware text transformation to be programmatically integrated into agentic workflows, developer tooling, and interactive AI systems.

\section{Privacy Evaluation}
In this section, we evaluate the anonymization performance quantitatively.
% from a privacy perspective, focusing on the ability of modern transformer-based Named Entity Recognition (NER) models to reliably detect privacy-bearing information which is a prerequisite for safe data sharing, de-identification, and downstream Natural Language Processing (NLP) applications on sensitive corpora.
To demonstrate the robustness and domain generalization of Redakto's anonymization capabilities, we performed experiments on three datasets from distinct domains: CodE Alltag (emails)\footnote{\url{https://github.com/codealltag/CodEAlltag}}, GraSCCo (clinical texts)\footnote{\url{https://zenodo.org/records/15747389}} and LER (legal documents)\footnote{\url{https://huggingface.co/datasets/elenanereiss/german-ler}}.
In summary the transformer models optimized for the Redakto demonstrator reach macro F1 $\approx$ 0.95 for PII detection when trained on domains with abundant annotated data (e.g., CodE Alltag, LER). For the low-resource clinical GraSCCo corpus performance is lower, in line with prior work such as \cite{kn:arzideh2025transformer}, which reports up to 0.95 macro F1 using thousands of in-domain clinical documents from University Hospital Essen (with neither data nor models publicly available), highlighting that data availability, annotation quality, and entity coverage - rather than model architecture - are the main bottlenecks for privacy-aware named entity recognition (NER).

\subsection{CodE Alltag - German Email Text}

Building upon the prior work of PII detection and pseudonymization of German-language text using the CodE Alltag corpus \cite{kn:krieg-holz-etal-2016-code,kn:eder-etal-2019-de,kn:eder2020code,kn:eder-etal-2022-beste,kn:saha2025pseudonymization}, we extend our experiments with a substantially larger sample size and a broader range of transformer-based model architectures for personally identifiable information (PII) detection. The full CodE Alltag corpus comprises approximately 1.5 million email messages, representing diverse instances of naturally occurring PII in informal and semi-formal communication. To obtain robust and reliable performance estimates, we adopt a 5-fold cross-validation setup over a subset of 175 thousand email texts using a 60/20/20 train–development–test split.
Further details on the training setup, model configurations, evaluation protocol,
and complete experimental results are provided in the project repository.\footnote{\url{https://github.com/calgo-lab/redakt-codealltag}}

\begin{figure}[H]
\centering
\begin{minipage}{\linewidth}
\centering
\begin{table}[H]
\small
\centering
\setlength{\tabcolsep}{13pt}
\caption{\centering PII-detection performance on CodE Alltag}
\begin{tabular}{| r | r | r | r |}
\hline
Model & Prec & Rec & F1 \\ \hline\hline
xlm-roberta-large & \cellcolor{gray!30}0.9442 ±0.0 & 0.9358 ±0.0 & 0.9399 ±0.0 \\ \hline
gelectra-large & 0.9433 ±0.0 & \cellcolor{gray!30}0.9371 ±0.0 & \cellcolor{gray!30}0.9401 ±0.0 \\ \hline
bert-base-german-cased & 0.9251 ±0.0 & 0.9160 ±0.0 & 0.9204 ±0.0 \\ \hline
\end{tabular}
\label{tab:pii_detection_performance_on_codealltag}
\end{table}
\end{minipage}
\\[10pt]
\begin{minipage}[c]{0.46\linewidth}
\centering
\includegraphics[width=\linewidth]{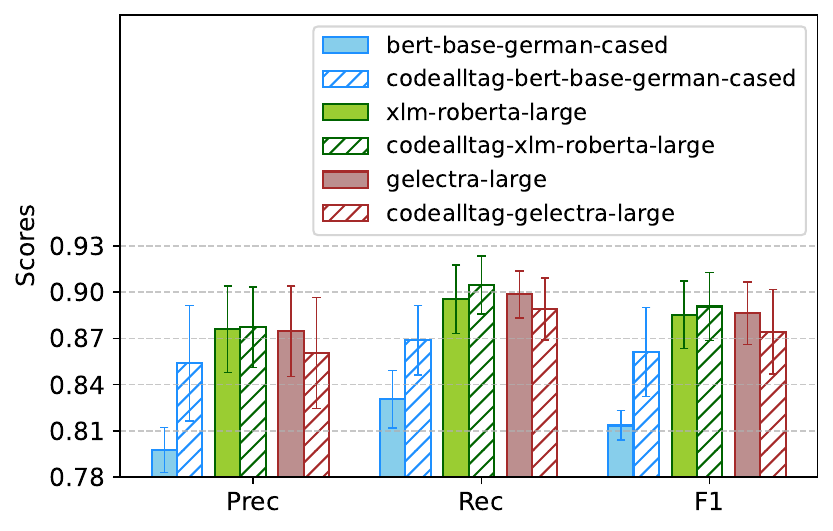}
\caption{\centering PII-detection performance on GraSCCo}
\label{fig:pii_detection_performance_on_grascco}
\end{minipage}
\begin{minipage}[c]{0.52\linewidth}
\centering
\begin{table}[H]
\small
\centering
\setlength{\tabcolsep}{4pt}
\caption{\centering PII-detection performance on LER}
\begin{tabular}{| r | r | r | r |}
\hline
Model & Prec & Rec & F1 \\ \hline\hline
xlm-roberta-large & 0.94 & 0.95 & 0.94 \\ \hline
gelectra-large & \cellcolor{gray!30}0.95  & \cellcolor{gray!30}0.96 & \cellcolor{gray!30}0.95 \\ \hline
bert-base-german-cased & 0.95 & 0.94 & 0.95 \\ \hline
\end{tabular}
\label{tab:pii_detection_performance_on_ler}
\end{table}
\end{minipage}
\end{figure}

\noindent We fine-tune and compare three transformer-based  models, evaluating their effectiveness in detecting 14 categories of PII in German email text. 
\autoref{tab:pii_detection_performance_on_codealltag} shows comparisons of macro averaged performance on the test sets with mean and standard deviation across the five cross-validation folds.

\subsection{GraSCCo - German Clinical Text}
% Strict privacy regulations and institutional barriers make it difficult to release real clinical text, even in de-identified form.
To study PII detection in German clinical documents, we make use of the Graz Synthetic Clinical Corpus (GraSCCo)~\cite{kn:modersohn2022grascco,kn:lohr2024deidentifying}.
% , a collection of artificially generated German-language clinical summaries designed to support clinical NLP research.
% The corpus was later annotated \cite{kn:lohr2024deidentifying} with Protected Health Information (PHI) entities using the INCEpTION\footnote{\url{https://inception-project.github.io/}} annotation platform.
% While GraSCCo provides a valuable starting point, it remains 
% a relatively small resource with 1,439 annotated PHI entities across 19 PII label types.
Given the small size of the GraSCCo corpus (1,439 annotated private entities across 19 PII label types), we explored transfer learning using PII detection model checkpoints previously fine-tuned on the CodE Alltag corpus. 
The task-specific classification heads were replaced to predict GraSCCo PII labels, and the models were further fine-tuned on the clinical corpus. \autoref{fig:pii_detection_performance_on_grascco} summarizes the micro-averaged evaluation results for all models on the GraSCCo test splits of five cross-validation folds. 
We also list here the GitHub repository\footnote{\url{https://github.com/calgo-lab/redakt-grascco}} and other resources \footnote{\url{https://grascco.demo.calgo-lab.de}} for more information.

\subsection{LER - German Legal Text}
To broaden the privacy-oriented evaluation beyond email and clinical text, we additionally fine-tuned the same three model architectures on the German Legal Entity Recognition (LER) dataset\cite{kn:leitner-etal-2020-dataset,kn:leitner2019finegrained}. Although LER is not a dedicated PII corpus, it contains a substantial proportion of entities that overlap with privacy-relevant categories, namely persons, locations, and organizations, which together account for 25.66\% of all annotated entities. The remaining 74.34\% correspond to types of entities specific to the legal-domain, such as legal norms, regulations, court decisions etc. This mixture allows us to assess how robustly models can identify classical personal identifiers when they appear alongside dense domain-specific terminology. We adopt the original train–development–test split provided with the dataset and do not introduce additional resampling or cross-validation. 
\autoref{tab:pii_detection_performance_on_ler} depicts the re-calculated macro averaged performance of the models for detecting entities of 12 PII labels present on test samples of LER dataset.

\section{Utility Evaluation}

We study the utility of anonymized texts by training models on original texts and evaluating classification performance on redacted texts as an extension of a prior work by \cite{kn:pal2024empirical} with two new tasks, medical intent classification and legal violation prediction.
Across these domains, we compare utility by replacing PII with semantic placeholders, random and generic masks to analyze how different anonymization strategies influence loss of task specific linguistic information, model robustness, and downstream utility. 
For additional experimental details and to facilitate reproducibility we refer readers to the project GitHub repository\footnote{\url{https://github.com/calgo-lab/redacted-text-utility}}.

\subsection{Medical Intent Classification}
For our first experiment, we use the Medical Intent Classification (MIC) dataset\footnote{\url{https://huggingface.co/datasets/DATEXIS/med_intent_classification}} introduced by \cite{kn:roehr2025wheredoeshurt} where the researchers study on physician intent trajectories in doctor–patient dialogues. 
The dataset is derived from the Ambient Clinical Intelligence Benchmark (ACI-Bench)\cite{kn:yim2023acibench} corpus and contains 5,541 physician turns annotated with 20 fine-grained intent taxonomy. 
Each sample consists of a single physician utterance, and the objective is to predict one or more associated medical intents, making this a multi-label classification problem.
We use a model\footnote{\url{https://huggingface.co/flair/ner-english-ontonotes-large}} finetuned with Flair~\cite{kn:akbik2019flair,kn:schweter2020flert} on OntoNotes~\cite{kn:weischedel2011ontonotes}, an English NER corpus\footnote{\url{https://catalog.ldc.upenn.edu/LDC2013T19}}, to identify and redact private entities (PERSON, DATE, GPE, ORG), with additional filtering to remove spurious detections.

\begin{figure}[H]
\centering
\begin{minipage}{\linewidth}
\centering
\begin{table}[H]
\small
\centering
\setlength{\tabcolsep}{5pt}
\caption{Examples of redaction strategies applied to a clinical text}
\begin{tabular}{p{0.38\linewidth} | p{0.56\linewidth}}
\hline
Version & Text \\ \hline\hline

Original 
& miss edwards is here for evaluation of facial pain this is a 54-year-old female \\ \hline

Semantic Label Masking 
& miss [PERSON] is here for evaluation of facial pain this is a [DATE] female \\ \hline

Random Masking 
& miss lhyZXSX is here for evaluation of facial pain this is a vejE4fPRUxkG female \\ \hline

Generic Masking 
& miss XXXX is here for evaluation of facial pain this is a XXXX female \\ \hline

\end{tabular}
\label{tab:examples_of_redaction_strategies_applied_to_a_clinical_text}
\end{table}
\end{minipage}
\\
\begin{minipage}[c]{0.50\linewidth}
\centering
\includegraphics[width=\linewidth]{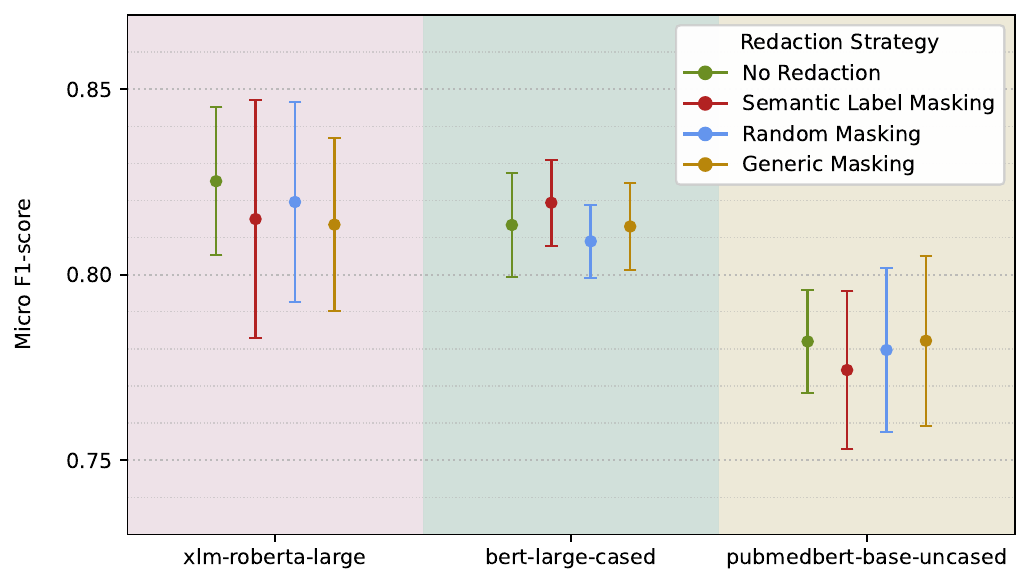}
\caption{\centering Medical Intent classification on anonymized texts}
\label{fig:medical_intent_classification_on_anonymized_texts}
\end{minipage}
\begin{minipage}[c]{0.48\linewidth}
\centering
\begin{table}[H]
\centering
\setlength{\tabcolsep}{3.2pt}
\small
\caption{\centering Binary Violation prediction on anonymized texts}
\begin{tabular}{| p{45pt} | p{30pt} | p{30pt} | p{35pt} |}
\hline
Redaction Strategy      & xlm-roberta-large  & bert-large-cased  & electra-large-discri. \\ \hline\hline
No Redaction            & 0.86 ±0.01         & 0.85 ±0.01        & 0.84 ±0.01                  \\ \hline
Semantic Label Masking  & 0.86 ±0.01         & 0.84 ±0.01        & 0.82 ±0.02                  \\ \hline
Random Masking          & 0.85 ±0.01         & 0.82 ±0.02        & 0.82 ±0.01                  \\ \hline
Generic Masking         & 0.84 ±0.02         & 0.85 ±0.01        & 0.82 ±0.02                  \\ \hline
\end{tabular}
\label{tab:binary_violation_prediction_on_anonymized_texts}
\end{table}
\end{minipage}
\end{figure}

\noindent We fine-tune three different transformer models for the multi-label intent classification downstream task and report performance metrics for each redaction (\autoref{tab:examples_of_redaction_strategies_applied_to_a_clinical_text}) condition in \autoref{fig:medical_intent_classification_on_anonymized_texts} where error bars denote performance variability across the five cross-validation folds.

\subsection{Binary Violation Prediction}
\label{sec:sec_binary_violation_prediction}
We conduct a second utility evaluation on the European Court of Human Rights (ECHR) dataset\footnote{\url{https://huggingface.co/datasets/glnmario/ECHR}} introduced by \cite{kn:chalkidis2019neural}.
% which contains about 11.5k cases with long factual descriptions. 
In line with the original work, we restrict our experiments to the Binary Violation Prediction (BVP) task, i.e., determining whether any human rights article of the European Convention on Human Rights was violated.
%In contrast to the medical intent dataset, ECHR documents are substantially longer and contain a higher density of private entities.
% , making this corpus both token-heavy and entity-heavy.
We utilize an experimental setup similar to that used for the MIC task (same redaction model and strategies).
Three transformer-based document classifiers are fine-tuned using the Flair framework with long-sequence support.
\autoref{tab:binary_violation_prediction_on_anonymized_texts} shows macro averaged performance of these models for different redaction strategies evaluated on the test set to measure the utility impact of privacy-driven sanitization. Our results demonstrate that redaction does not impact utility substantially. The least effect of redaction is observed with semantic label masking. We present a more in-depth analysis, controlling for the number of PII entities in a text for this task in Appendix~\ref{appendix:additional_experimental_results}. In the majority of the experimental conditions investigated, redaction appears to have limited impact on the utility of the anonymized texts.

\section{Conclusion}
% In this work we provide technology to enable end-users, engineers and researchers to leverage LLMs in a privacy preserving manner. 
Our Redakto demonstrator provides a web application, REST APIs and MCP hooks for direct usage inside an LLM. In contrast to other implementations of redaction tools for LLM usage we combine the implementation with comprehensive evaluations on redaction performance and utility of redacted texts. Empirical evaluations demonstrate not only that Redakto reliably anonymizes texts across a variety of domains, our results also show that the utility of the anonymized texts is not  impacted substantially in the majority of cases, even when controlling for the number of PII entities in a given text. These findings highlight the potential of Redakto for researchers, practitioners and every-day usage.

\section*{Acknowledgements}
This research was supported by the German Federal Ministry of Research, Technology and Space grant numbers 16SV8857, by the Einstein Center Digital Future, Berlin, and by the German Research Foundation (DFG) - Project number: 528483508 - FIP 12. 
\bibliographystyle{splncs04}
\bibliography{wipeout26}

\clearpage

\appendix
\renewcommand{\theHsection}{appendix.\Alph{section}}
\renewcommand{\theHfigure}{appendix.\Alph{section}.\arabic{figure}}
\section{Additional Experimental Results}
\label{appendix:additional_experimental_results}
The ECHR dataset used for one of our utility evaluation tasks (\ref{sec:sec_binary_violation_prediction}) represents a particularly interesting setting for some additional experiments - after preprocessing and sample selection across all the test folds, documents contain on average approximately 2000 tokens and 86 private entities, with most entity-dense documents containing several hundred PII entities.
This makes the dataset well suited for evaluating whether and by how much extensive redaction affects downstream model performance.

% In \autoref{fig:macro_f1_xlm-roberta-large_no_redaction_vs_different_redaction_strategies}, we analyse the impact of redaction strategies on the utility of anonymized texts as measured in terms of Macro-F1 scores. Our results show that utility is not impacted by the redaction. Semantic label masking (\autoref{fig:macro_f1_xlm-roberta-large_no_redaction_vs_different_redaction_strategies}, top panel) appears to show the least impact on utility.

\begin{figure}
\centering
\begin{minipage}[c]{0.9\linewidth}
\centering
\includegraphics[width=\linewidth,height=0.23\textheight,keepaspectratio]{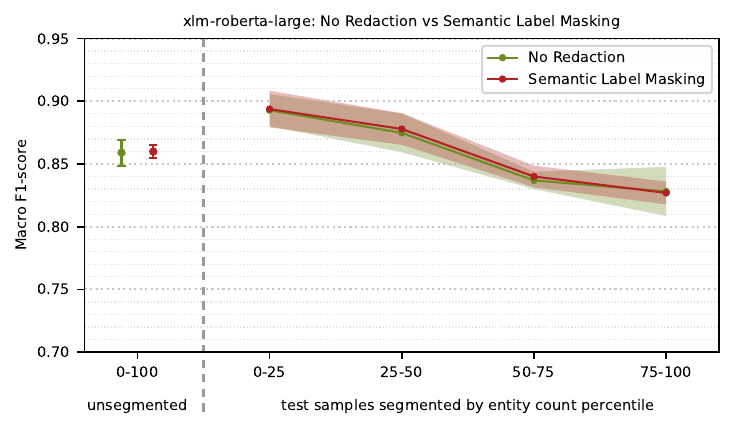}
\end{minipage}
\begin{minipage}[c]{0.98\linewidth}
\centering
\includegraphics[width=\linewidth,height=0.23\textheight,keepaspectratio]{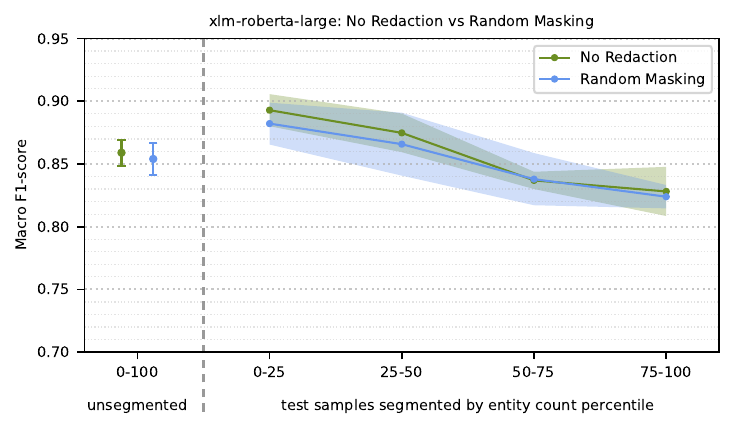}
\end{minipage}
\begin{minipage}[c]{0.98\linewidth}
\centering
\includegraphics[width=\linewidth,height=0.22\textheight,keepaspectratio]{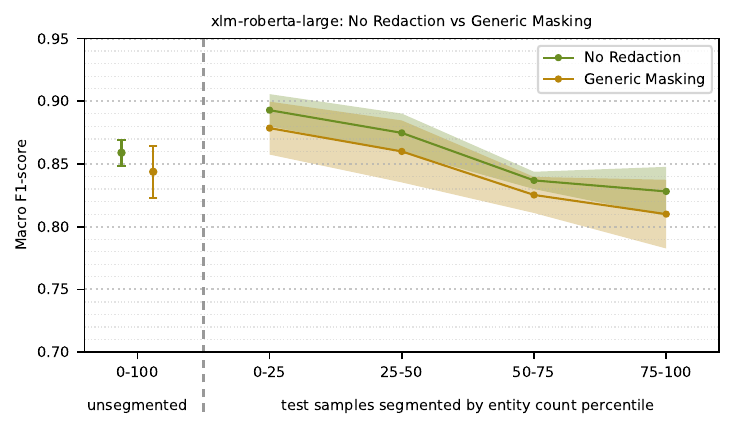}
\end{minipage}
\caption{\centering Entity density wise performance (Macro-F1) comparison for different redaction strategies (xlm-roberta-large)}
\label{fig:macro_f1_xlm-roberta-large_no_redaction_vs_different_redaction_strategies}
\end{figure}

\noindent To investigate the robustness of different redaction strategies under increasing amounts of sensitive information, we perform an additional analysis based on entity density.
For each cross-validation fold, test documents are ranked according to their number of detected entities and partitioned into four percentile ranges (0–25, 25–50, 50–75, and 75–100).
Classification performance is then evaluated separately for each subset for all fine-tuned binary violation prediction models and for each redaction strategy.

\begin{figure}
\centering
\begin{minipage}[c]{0.98\linewidth}
\centering
\includegraphics[width=\linewidth,height=0.23\textheight,keepaspectratio]{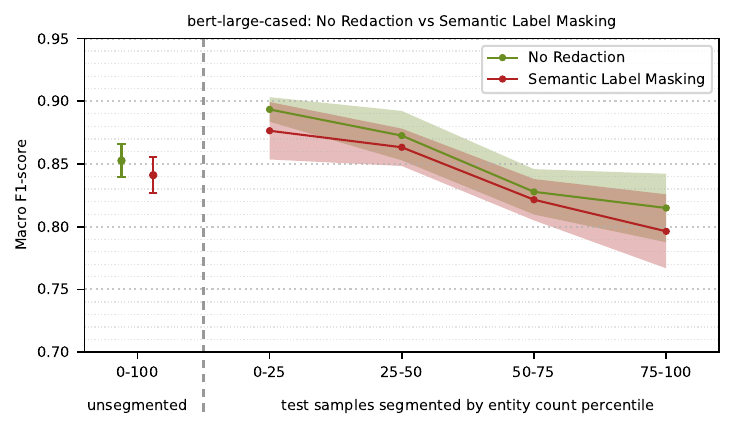}
\end{minipage}
\begin{minipage}[c]{0.98\linewidth}
\centering
\includegraphics[width=\linewidth,height=0.23\textheight,keepaspectratio]{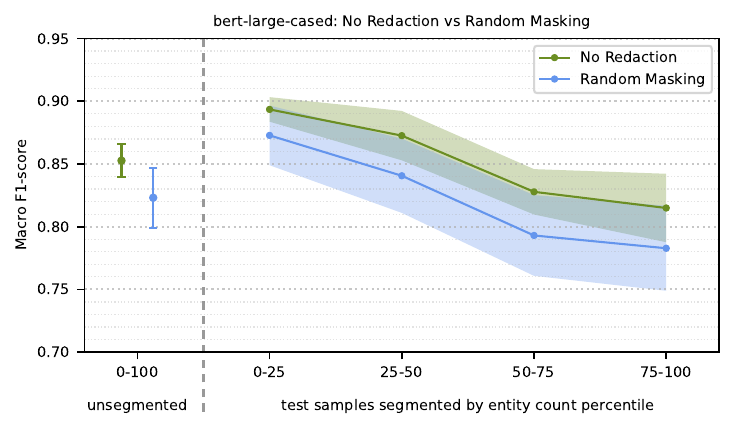}
\end{minipage}
\begin{minipage}[c]{0.98\linewidth}
\centering
\includegraphics[width=\linewidth,height=0.23\textheight,keepaspectratio]{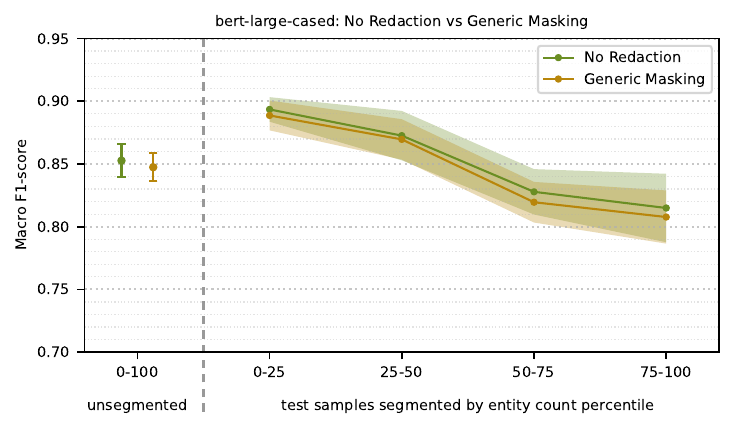}
\end{minipage}
\caption{\centering Entity density wise performance (Macro-F1) comparison for different redaction strategies (bert-large-cased)}
\label{fig:macro_f1_bert-large-cased_no_redaction_vs_different_redaction_strategies}
\end{figure}

\noindent This experiment enables a fine-grained assessment of utility preservation under progressively more intense redaction scenarios.
Since higher entity-count percentiles require substantially larger portions of a document to be transformed, performance trends across percentile ranges provide insight into whether redaction introduces additional degradation beyond the inherent difficulty of processing longer documents.

\begin{figure}[H]
\centering
\begin{minipage}[c]{0.95\linewidth}
\centering
\includegraphics[width=\linewidth,height=0.24\textheight,keepaspectratio]{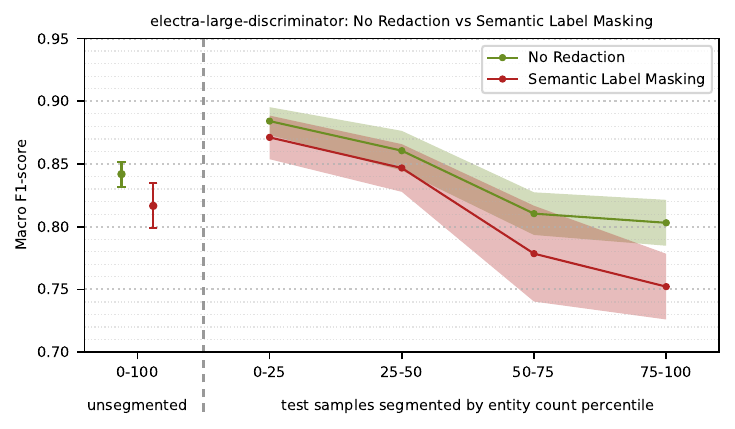}
\end{minipage}
\begin{minipage}[c]{0.95\linewidth}
\centering
\includegraphics[width=\linewidth,height=0.24\textheight,keepaspectratio]{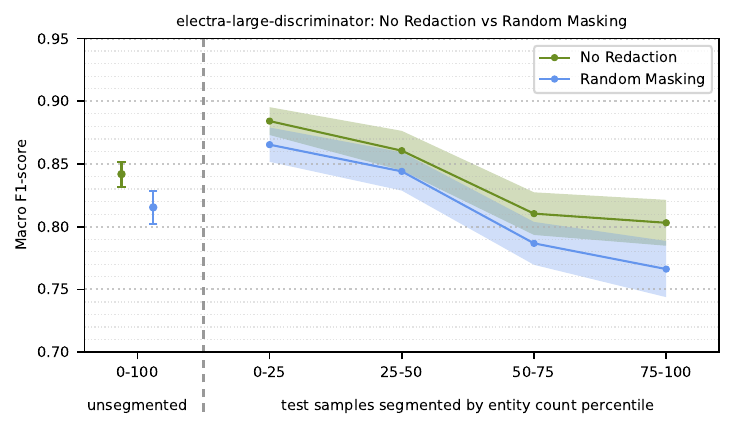}
\end{minipage}
\begin{minipage}[c]{0.95\linewidth}
\centering
\includegraphics[width=\linewidth,height=0.24\textheight,keepaspectratio]{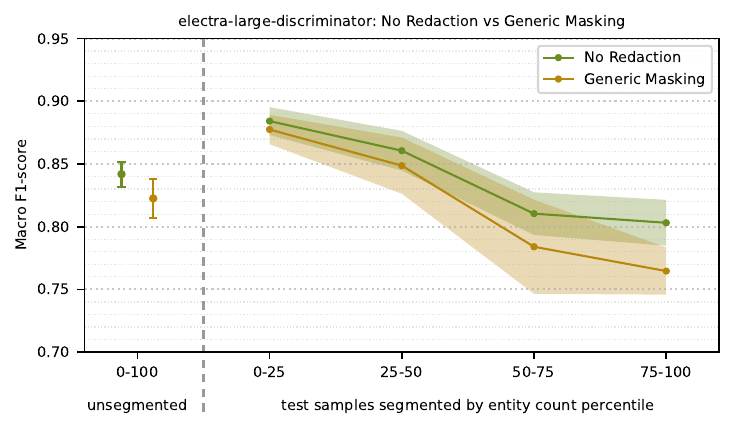}
\end{minipage}
\caption{\centering Entity density wise performance (Macro-F1) comparison for different redaction strategies (electra-large-discriminator)}
\label{fig:macro_f1_electra-large-discriminator_no_redaction_vs_different_redaction_strategies}
\end{figure}

\noindent We present the results of this experiment with \autoref{fig:macro_f1_xlm-roberta-large_no_redaction_vs_different_redaction_strategies}, \autoref{fig:macro_f1_bert-large-cased_no_redaction_vs_different_redaction_strategies} and \autoref{fig:macro_f1_electra-large-discriminator_no_redaction_vs_different_redaction_strategies}.
As entity density increases, documents also become substantially longer and more complex, resulting in a gradual performance decline across all models and redaction strategies.
However, the relative performance gap remains largely stable across density levels in comparison to the unredacted counterpart, suggesting that document complexity -- not redaction -- is the primary driver of degradation, with electra-large-discriminator showing slightly greater sensitivity that may be related to its replaced-token-detection pre-training objective.

\section{Redakto System Design}
\label{appendix:redakto_system_design}

\noindent Redakto is organized as a layered text-redaction system consisting of a
Streamlit-based interaction layer and a FastAPI backend.
Through the user interface, users select an entity set, label granularity, and model.
The interface then submits structured JSON requests to backend prediction endpoints
for either named entity recognition or pseudonym generation.

The backend validates incoming requests using Pydantic schemas and delegates
inference to the application-level prediction service.
This service dispatches requests according to the configured model type.
NER models return detected entity spans, whereas NER-PG models return entity spans together with generated pseudonyms and reconstructed pseudonymized text variants.

\begin{figure}[htbp]
    \centering
    \includegraphics[width=\textwidth]{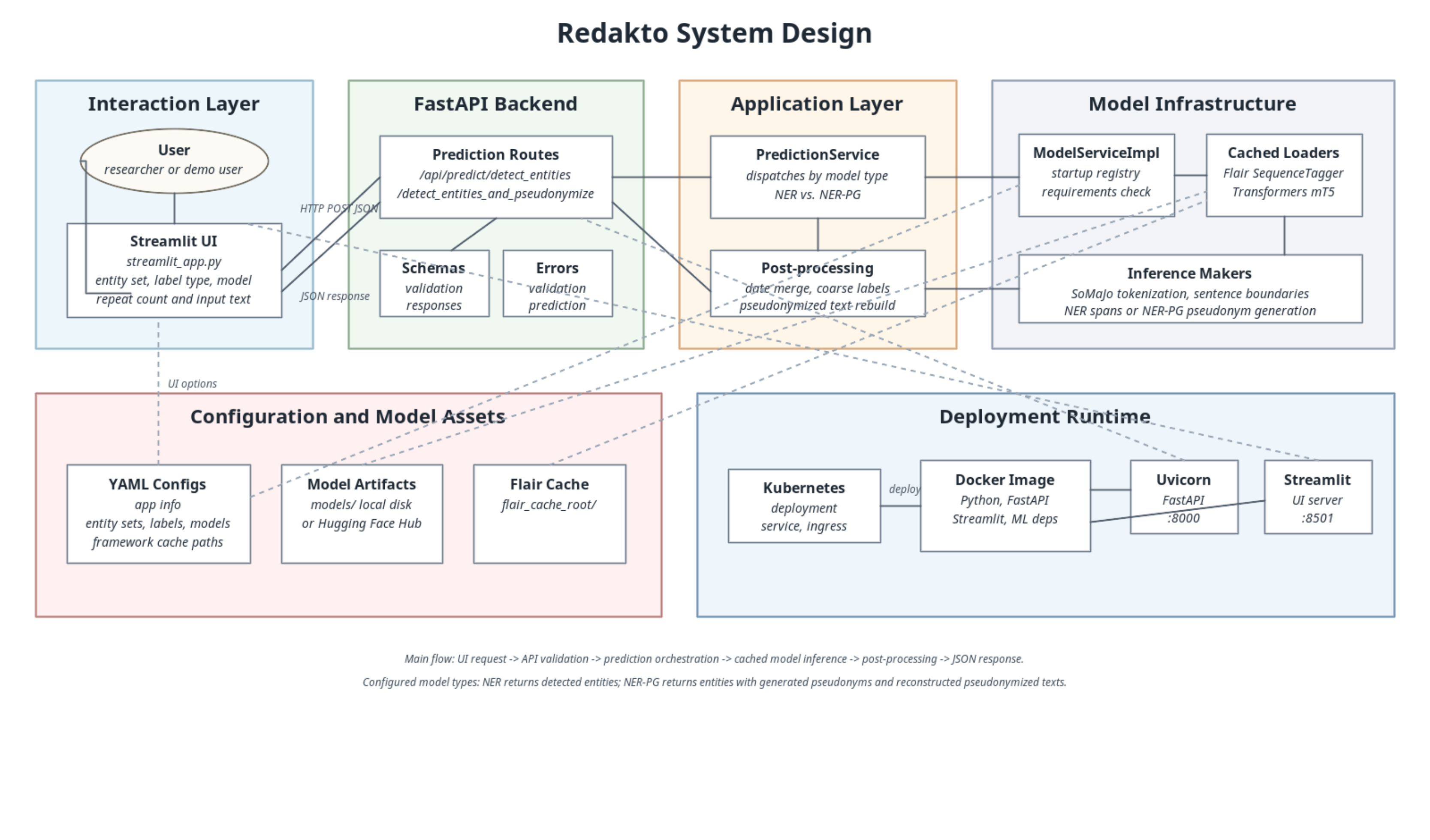}
    \caption{System design of Redakto}
    \label{fig:redakto_system_design}
\end{figure}

\noindent Model loading and inference are handled by a separate infrastructure layer.
At startup, the application builds a model registry from YAML configuration files
that define the available entity sets, labels, model metadata, loading
strategies, and framework paths.
Cached loaders support Flair sequence-tagging models and Transformer-based mT5 models.
The inference adapters perform SoMaJo tokenization, sentence boundary handling, and chunking for longer inputs.

After inference, Redakto applies task-specific post-processing steps, including
the merging of adjacent date entities, optional mapping from fine-grained to
coarse-grained labels, and reconstruction of pseudonymized text.
The system is containerized with Docker and includes Kubernetes manifests for deploying both
the user interface and the API services.


% LNCS Homepage, \url{http://www.springer.com/lncs}, last accessed 2023/10/25
% \end{thebibliography}
\end{document}